\documentclass[journal]{IEEEtran}
\usepackage{graphics}
\usepackage{graphicx}
\usepackage{caption}
\usepackage{amsfonts}
\usepackage{multirow}
\usepackage{gensymb}
\usepackage{subfigure}
\usepackage{palatino}
\usepackage{tikz}
\usepackage{latexsym}
\usetikzlibrary{shapes.geometric, arrows}

\begin{document}

\title{PointHop: An Explainable Machine Learning Method for Point Cloud Classification}

\author{Min Zhang, ~\IEEEmembership{Student member,~IEEE,} 
Haoxuan You, Pranav Kadam, ~\IEEEmembership{Student member,~IEEE,} 
Shan Liu, ~\IEEEmembership{Senior member,~IEEE,} and C.-C. Jay Kuo, ~\IEEEmembership{Fellow,~IEEE}
\thanks{Min Zhang is with the Department of Electrical and Computer
Engineering, Viterbi School of Engineering, University of Southern
California, CA, 90007 USA (e-mail: zhan980@usc.edu).}
\thanks{Haoxuan You is with the Department of Computer Science, Columbia
University, NY, 10027 USA (email: hy2612@columbia.edu).}
\thanks{Pranav Kadam is with the Department of Electrical and Computer
Engineering, Viterbi School of Engineering, University of Southern
California, CA, 90007 USA (e-mail: pranavka@usc.edu).}
\thanks{Shan Liu is with Tencent Media Lab, Tencent America, 2747 Park
Blvd, Palo Alto, CA, 94306 USA (email: shanl@tencent.com).}
\thanks{C.-C. Jay Kuo is with the Media Communications Lab of the
Department of Electrical and Computer Engineering, University of
Southern California, CA, 90007 USA (e-mail: cckuo@sipi.usc.edu).}
}
\maketitle

\begin{abstract} 
An explainable machine learning method for point cloud classification, 
called the PointHop method, is proposed in this work. The PointHop 
method consists of two stages: 1) local-to-global attribute building 
through iterative one-hop information exchange, and 2) classification and 
ensembles. In the attribute building stage, we address the problem of 
unordered point cloud data using a space partitioning procedure and 
developing a robust descriptor that characterizes the relationship between a
point and its one-hop neighbor in a PointHop unit. When we put multiple
PointHop units in cascade, the attributes of a point will grow by taking its
relationship with one-hop neighbor points into account iteratively.
Furthermore, to control the rapid dimension growth of the attribute vector
associated with a point, we use the Saab transform to reduce the attribute
dimension in each PointHop unit. In the classification and ensemble stage, 
we feed the feature vector obtained from multiple PointHop units to a
classifier. We explore ensemble methods to improve the classification
performance furthermore. It is shown by experimental results that the 
PointHop method offers classification performance that is comparable with
state-of-the-art methods while demanding much lower training complexity. 
\end{abstract}


\IEEEpeerreviewmaketitle

\begin{figure}[htpb]
\centering
\includegraphics[width=3.5in]{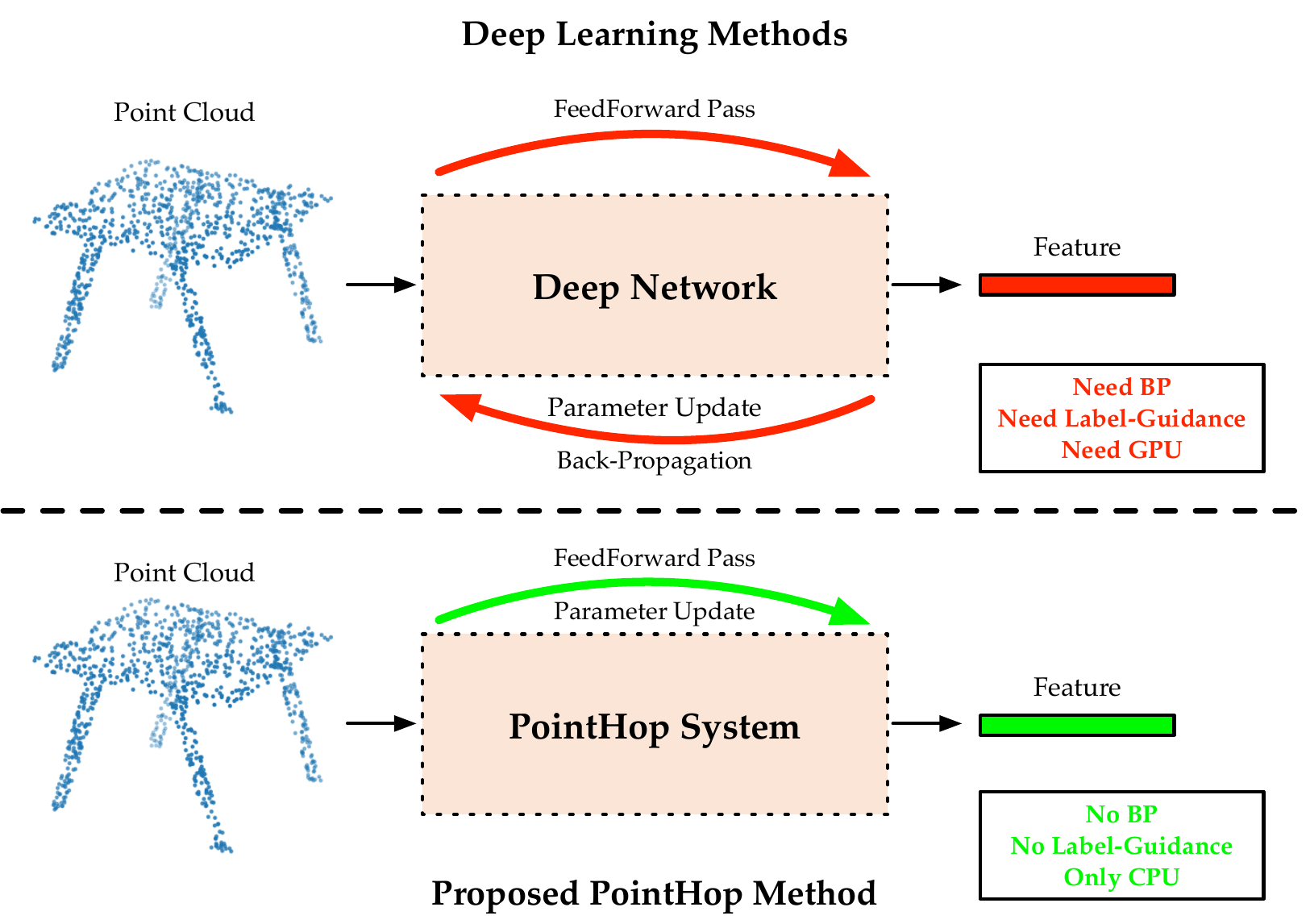}
\caption{Comparison of existing deep learning methods and the proposed
PointHop method. Top: Point cloud data are fed into deep neural networks
in the feedforward pass and errors are propagated in the backward
direction. This process is conducted iteratively until convergence.
Labels are needed to update all model parameters. Bottom: Point cloud
data are fed into the PointHop system to build and extract features in
one fully explainable feedforward pass. No labels are needed in the feature
extraction stage (i.e. unsupervised feature learning). The whole training of
PointHop can be efficiently performed on a single CPU at much lower
complexity than deep-learning-based methods.} \label{fig:my_label-1}
\end{figure}

\section{Introduction}\label{sec:introduction}

\IEEEPARstart{T}{hree} dimensional (3D) object classification and
recognition is one of the fundamental problems in multimedia and
computer vision. 3D objects can be represented in different forms, one
of which is the point cloud model. Point cloud models are popular due to
easy access and complete description in the 3D space. It has been
widely studied in the research community. Most state-of-the-art methods
extract point cloud features by building deep neural networks and using
backpropagation to update model parameters iteratively. However, deep
networks are difficult to interpret. Their training cost is so high that
the GPU resource is inevitable. Furthermore, the requirement of
extensive data labeling adds another burden. All these concerns impede
reliable and flexible applications of the deep learning solution in 3D
vision. To address these issues, we propose an explainable machine
learning method, called the PointHop method, for point cloud
classification in this work. 

A 3D object can be represented in one of the following four forms: a
voxel grid, a 3D mesh, multi-view camera projections, and a point cloud.
With proliferation of deep learning, many deep networks have been
designed to process different representations, e.g.,
\cite{qi2016volumetric, you2018pvnet, riegler2017octnet, papadakis2010panorama}. Voxel grids use occupancy cubes to describe 3D
shapes. Some methods \cite{maturana2015voxnet, brock2016generative}
extend the 2D convolution to the 3D convolution to process the 3D
spatial data. Multi-view image data are captured by a set of cameras
from different angles. A weight-shared 2D convolutional neural network
(CNN) is applied to each view, and results from different views are
fused by a view aggregation operation in \cite{su2015multi,feng2018gvcnn}. Feng {\em et al.} \cite{feng2018gvcnn} proposed a
group-view CNN (GVCNN) for 3D objects, where discriminability of each
view is learned and used in the 3D representation. The 3D mesh data
contains a collection of vertices, edges and faces. The MeshNet
\cite{feng2018meshnet} treats faces of a mesh as the basic unit and
extracts their spatial and structural features individually to offer the
final semantic representation. By considering multimodal data, Zhang
{\em et al.} \cite{zhang2018inductive} proposed a hypergraph-based
inductive learning method to recognize 3D objects, where complex
correlation of multimodal 3D representations is explored. 

A point cloud is represented by a set of points in the 3D coordinates.
Among the above-mentioned four forms, point clouds are easiest to
acquire since they can be directly obtained by the LiDAR and the RGB-D
sensors. Additionally, the point cloud data has more complete
description of 3D objects than other forms. Because of these properties,
point clouds are deployed in various applications ranging from 3D
environment analysis \cite{landrieu2018large, angelina2018pointnetvlad}
to autonomous driving \cite{yang2018hdnet, yang2018pixor, lang2019pointpillars}. They have attracted increasing attention from the
research community in recent years.  

State-of-the-art point cloud classification and segmentation methods are
based on deep neural networks. Points in a point cloud are irregular and
unordered so they cannot be easily handled by regular 2D CNNs. To
address this problem, PointNet \cite{qi2017pointnet} uses multi-layer
perceptrons (MLPs) to extract features for each point separately. Then,
it is followed by a symmetric function to accumulate all point features.
Subsequent methods, including \cite{qi2017pointnet++, wang2018dynamic, shen2018mining}, focus on effectively processing the information of
neighboring points jointly rather than individually. PointNet++
\cite{qi2017pointnet++} utilizes the PointNet in sampled local regions
and aggregates features hierarchically. DGCNN \cite{wang2018dynamic}
builds dynamic connections among points in their feature level and
updates point features based on their neighboring points in the feature
space. 

Although deep-learning-based methods provide good classification
performance, their working principle is not transparent. Furthermore,
they demand huge computational resources (e.g., long training time even
with GPUs). Since it is challenging to deploy them in mobile or terminal
devices, their applicability to real world problems is hindered. To
address these shortcomings, we propose a new and explainable learning
method, called the PointHop method, for point cloud data recognition.
PointHop is mathematically transparent. We compare PointHop with
deep-learning-based methods in Fig. \ref{fig:my_label-1}. PointHop
requires only one forward pass to learn parameters of the system.
Furthermore, its feature extraction is an unsupervised procedure since
no class labels are needed in this stage. 

The PointHop method consists of two stages: 1) local-to-global attribute
building through iterative one-hop information exchange, and 2)
classification and ensembles. In the attribute building stage, we
address the problem of unordered point cloud data using a space
partitioning procedure and developing an effective and robust descriptor
that characterizes the relationship between a point and its one-hop
neighbor in a PointHop unit. 

When we put multiple PointHop units in cascade, the attributes of a
point will grow by taking its relationship with one-hop neighbor points
into account iteratively. Furthermore, to control the rapid dimension
growth of the attribute vector associated with a point, we use the Saab
transform to reduce the attribute dimension in each PointHop unit. In
the classification and ensemble stage, we feed the feature vector
obtained from multiple PointHop units to a classifier, such as the
support vector machine (SVM) classifier \cite{cortes1995support} and the
random forest (RF) classifier \cite{breiman2001random} to get
classification result. Furthermore, we explore ensemble methods to
improve the final classification performance. Extensive experiments are
conducted on the ModelNet40 dataset to evaluate the performance of the
PointHop method. We also compare PointHop with state-of-the-art deep
learning methods. It is observed that PointHop can achieve comparable
performance on 3D shape classification task with much lower training
complexity. For example, the training process takes of PointHop less
than 20 minutes with CPU while the training of deep learning methods
takes several hours even with GPU. 

The rest of this paper is organized as follows. Related work is reviewed
in Section \ref{sec:review}. Details of the proposed PointHop method are
presented in Section \ref{sec:pointhop}. Experimental results are shown
in Section \ref{sec:experiments}. Finally, concluding remarks are given
in Section \ref{sec:conclusion}. 

\section{Review of Related Work}\label{sec:review}

\subsection{Feedforward-designed CNNs (FF-CNNs)}

Deep learning is a black-box tool while its training cost is extremely
high. To unveil its mystery and reduce its complexity, a sequence of
research work has been conducted by Professor Kuo and his students at
the University of Southern California in the last five years, including
\cite{kuo2016understanding, chen2018saak, kuo2017cnn, kuo2018data, chen2019ensembles, kuo2019interpretable}. These prior arts lay the
foundation for this work.  

Specifically, Kuo pointed out the sign confusion problem arising from
the cascade of hidden layers in CNNs and argued the need of nonlinear
activation to eliminate this problem in \cite{kuo2016understanding}.
Furthermore, Kuo \cite{kuo2018data} interpreted the all filters in one
convolutional layer form a subspace so that each convolutional layer
corresponds to a subspace approximation to the input. However, the
analysis of subspace approximation is still complicated due to the
existence of nonlinear activation. It is desired to solve the sign
confusion problem with other means. The Saak transform
\cite{kuo2018data, chen2018saak} and the Saab transform
\cite{kuo2019interpretable} were proposed to achieve two objectives
simultaneously; namely, avoiding sign confusion and preserving the
subspace spanned by the filters fully. 

One important advantage of the Saak and the Saab transforms is that
their transform kernels (or filters) can be mathematically derived using 
the principal component analysis (PCA) \cite{wold1987principal}. 
Multi-stage Saab and Saak filters can be derived in an unsupervised and
feedforward manner without backpropagation. Generally speaking, the Saab
transform is more advantageous than the Saak transform since the number of
Saab filters is only one half of the Saak filters. Besides interpreting the
cascade of convolutional layers as a sequence of approximating
spatial-spectral subspaces, Kuo {\em et al.} \cite{kuo2019interpretable}
explained the fully connected layers as a sequence of ``label-guided
least-squared regression" processes. As a result, one can determine all model
parameters of CNNs in a feedforward one-pass fashion. It is called the
feedforward-designed CNNs (FF-CNNs). No backpropagation is applied in this
design at all. More recently, an ensemble scheme was introduced in
\cite{chen2019ensembles} to enhance the performance of FF-CNNs. FF-CNNs was
only tested on the MNIST and the CIFAR-10 datasets in
\cite{kuo2019interpretable}. It is not trivial to generalize it to the point
cloud classification problem since points in a point cloud are irregular and
unordered. 

\subsection{Point Cloud Processing Methods}

A point cloud is represented by a set of points with 3D coordinates
$\{x,y,z\}$. It is the most straightforward format for 3D object
representation since it can be acquired by the LiDAR and the RGB-D
sensors directly. Point clouds have drawn a lot of attention since they
have a wide range of applications ranging from AR/VR to autonomous
driving. Extracting features of point clouds effectively is a key step
to 3D object recognition. 

Traditionally, point cloud features are handcrafted for specific tasks.
The statistical attributes are encoded into point features, which are
often invariant under shape transformation. Kernel signature methods
were used to model intrinsic local structures in \cite{sun2009concise, bronstein2010scale, aubry2011wave}. The point feature histogram was
introduced in \cite{rusu2008aligning} for point cloud registration. It
was proposed in \cite{chen2003visual} to project 3D models into
different views for retrieval. Multiple features can be combined to meet
the need of several tasks. 

With the advancement of deep learning, deep networks have been employed
for point cloud classification. The PointNet \cite{qi2017pointnet} used
deep neural networks to process point clouds with a spatial transform
network and a symmetry function so as to achieve permutation
invariance. On the other hand, the local geometric information is vital
to 3D object description. This is however ignored by PointNet.
Effective utilization of the local information became the focus of
recent deep learning work on this topic. For instance, PointNet++
\cite{qi2017pointnet++} applied the PointNet structure in local point
sets with different resolutions and, then, accumulated local features in
a hierarchical architecture. The PointCNN \cite{li2018pointcnn} used the
$\chi$-Conv to aggregate features in each local pitch and adopted a
hierarchical network structure similar to typical CNNs. As to 3D object
detection, the Frustum-PointNet \cite{qi2018frustum} converted 2D
detection results into 3D frustums and, then, employed the PointNet
blocks to segment out proposals as well as estimate 3D locations. The
VoxelNet \cite{zhou2018voxelnet} partitioned an outdoor scene into voxel
grids, where inside points of each cube were gathered together to form
regional features. Finally, the 3D convolution was used to get 3D
proposals. However, the training of deep networks is computationally
expensive, which imposes severe constraints on their applicability on
mobile and/or terminal devices. 

\section{Proposed PointHop System}\label{sec:pointhop}

\begin{figure}[htpb]
\centering
\includegraphics[width=3.5in]{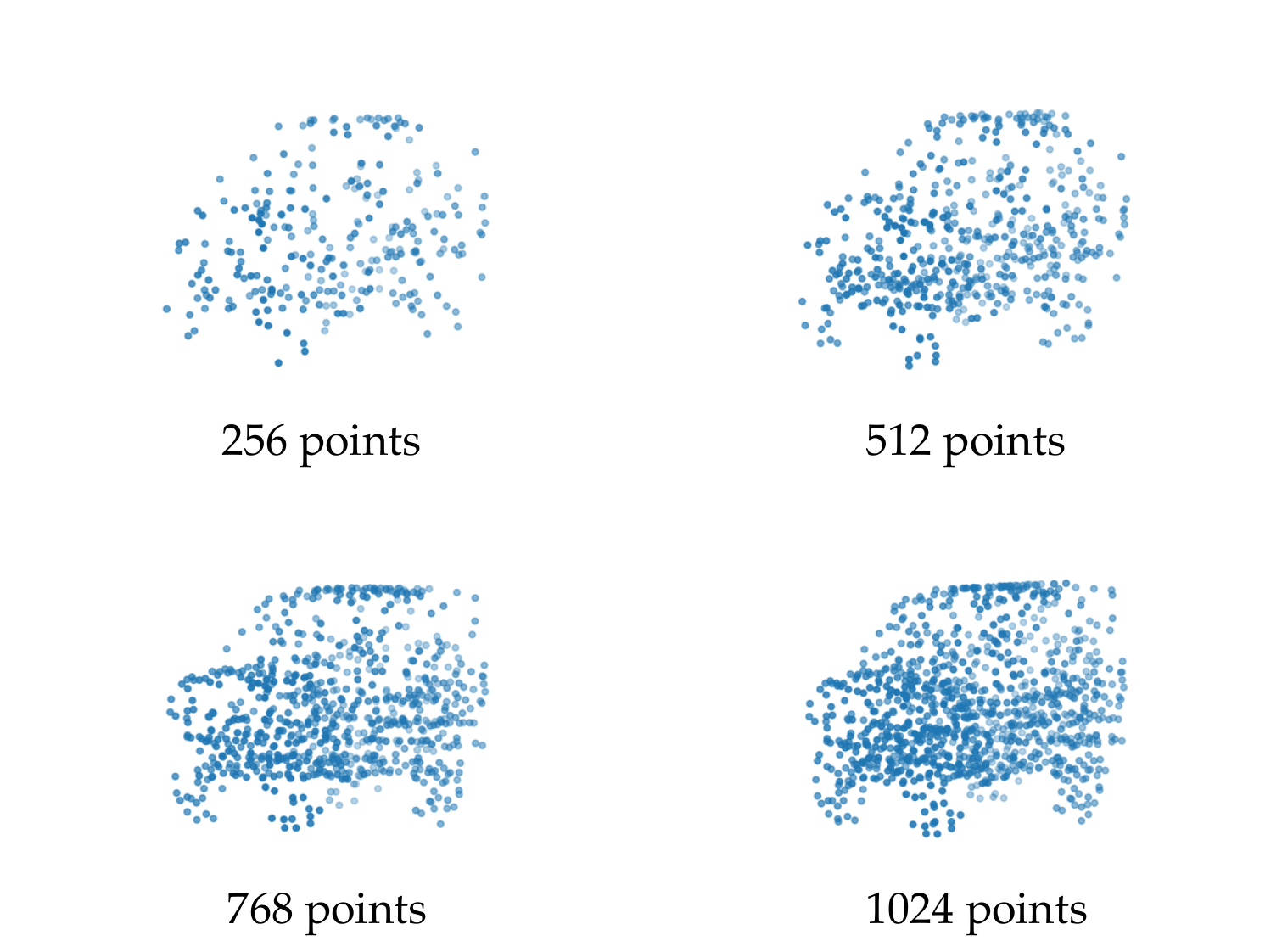}
\caption{Random sampling of a point cloud of 2,048 points into 
simplified models of (a) 256 points, (b) 512 points, (c) 768 points
and (d) 1,024 points. They are called the random dropout point
(DP) models and used as the input to the PointHop system.}
\label{fig:my_label-2}
\end{figure}

The source point cloud model typically contains a large number of points
of high density, and its processing is very time-consuming. We can apply
random sampling to reduce the number of points with little degradation
in classification performance. As shown in Fig. \ref{fig:my_label-2},
an exemplary point cloud model of 2,048 points is randomly sampled and
represented by four different point numbers. They are called the random
dropout point (DP) models. A model with more sampled points provides
higher representation accuracy at the cost of higher computational
complexity. We will use the DP model as the input to the proposed
PointHop system, and show the classification accuracy as a function of
the point numbers of a DP model in Sec. \ref{sec:experiments}. 

\begin{figure*}[htpb]
\centering
\includegraphics[width=7in]{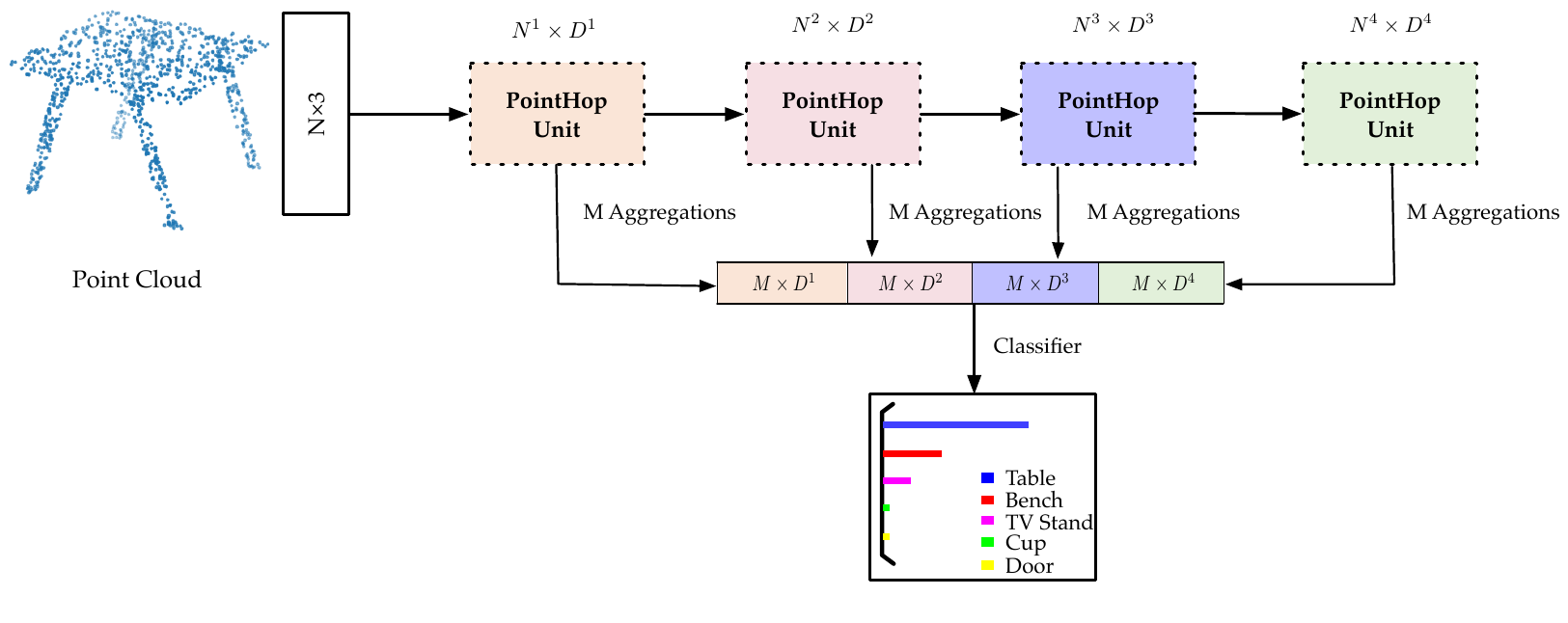}
\caption{An overview of the PointHop method. The input point cloud has
$N$ points with 3 coordinates $(x,y,z)$. It is fed into multiple PointHop
units in cascade and their outputs are aggregated by $M$ different schemes
to derive features. All features are cascaded for object classification.} \label{fig:my_label-3}
\end{figure*}

A point cloud of $N$ points is defined as ${\bf P}=\{{\bf p}_1, \cdots ,
{\bf p}_N\}$, where ${\bf p}_n \in \mathbb{R}^3$, $n=1,\cdots,N$. There
are two distinct properties of the point cloud data:
\begin{itemize}
\item unordered data in the 3D space \\
Being different from images where pixels are defined in a regular 2D
grid, a point cloud contains a set of points in the 3D space without a
specific order. 
\item disturbance in scanned points \\
For the same 3D object, Different point sets can be acquired with
uncertain position disturbance because of different scanning methods
applied to the surface of the same object or at different times using the
same scanning method. 
\end{itemize}
An overview of the proposed PointHop method is shown in Fig.
\ref{fig:my_label-3}. It takes point cloud, ${\bf P}$, as the input and
outputs the corresponding class label. It consists of two stages: 1)
local-to-global attribute building through multi-hop information
exchange, and 2) classification and ensembles. They will be elaborated
in Secs. \ref{subsec:feature} and \ref{subsec:decision}, respectively. 

\subsection{Local-to-Global Attribute Building}\label{subsec:feature}

\begin{figure*}[htpb]
\centering
\includegraphics[width=7.3in]{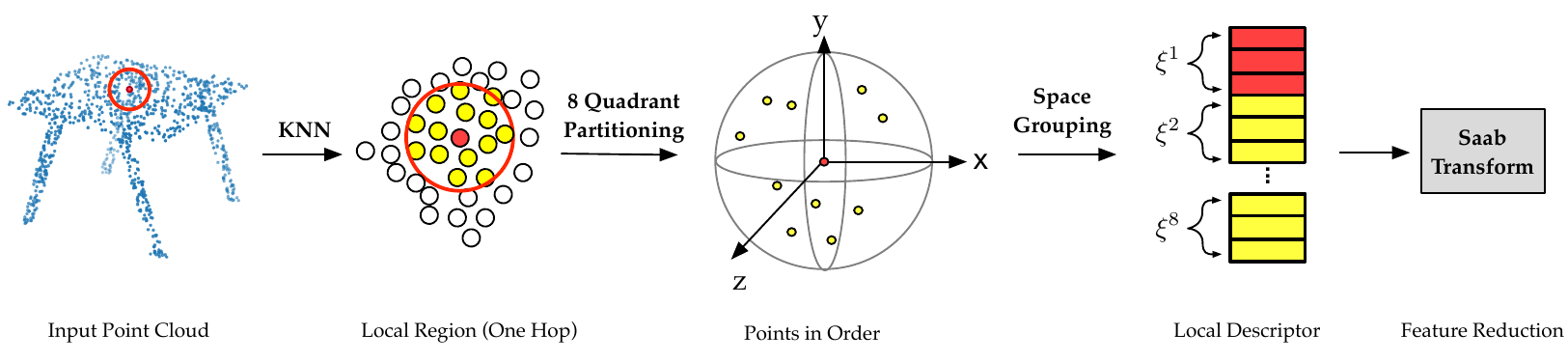}
\caption{Illustration of the PointHop Unit. The red point is the center 
point while the yellow points represent its $K$ nearest neighbor points.}
\label{fig:my_label-4}
\end{figure*}

In this subsection, we examine the evolution of attributes of a point in
${\bf P}$. Initially, the attributes of a point are its 3D coordinates.
Then, we use the attributes of a point and its neighboring points within
one-hop distance to build new attributes. Since the new attributes take
the relationship between multiple points into account, the dimension of
attributes grow. To control the rapid growth of the dimension, we apply
the Saab transform for dimension reduction. All these operations
are conducted inside a processing unit called the PointHop unit.

The PointHop unit is shown in Fig. \ref{fig:my_label-4}. It consists of
two modules: 
\begin{enumerate}
\item Constructing a local descriptor with attributes of one-hop neighbors \\
The construction takes issues of unordered 3D data and disturbance of
scanned points into account to ensure that the local descriptor is
robust. The attributes of a point evolve from a low dimensional vector
into a high dimensional one through this module. 
\item Using the Saab transform to reduce the dimension of the local descriptor \\
The Saab transform is used to reduce the dimension of the expanded attributes
so that the dimension grows at a slower rate.
\end{enumerate}

For each point in ${\bf P}$, ${\bf p}_c=(x_c, y_c, z_c)$, we search its $K$ 
nearest neighbor points in ${\bf P}$, including itself, where the distance is 
measured by the Euclidean norm. They form a local region: 
\begin{equation}
KNN({\bf p}_c) = \{{\bf p}_{c1}, \cdots , {\bf p}_{cK}\}, \quad {\bf p}_{c1}, 
\cdots, {\bf p}_{cK} \in {\bf P}.
\end{equation}
For each local region centered at ${\bf p}_c$, we treat ${\bf p}_c$ as
a new origin and partition it into eight quadrants $\xi^j$,
$j=1,\cdots,8$ based on the value of each coordinate (i.e., greater or
less than that of ${\bf p}_c$). 

We compute the centroid of attributes of points at each quadrant via
\begin{equation}
{\bf a}_{c}^{j} = \frac{1}{K_j}\sum_{i=1}^{K_j}t_{ci}^{j}{{\bf a}_{ci}}, 
\quad, j=1,\cdots,8,
\end{equation}
where ${\bf a}_{ci}$ is the attribute vector of point ${\bf p}_{ci}$ and 
\begin{equation}
t_{ci}^j=\left\{
\begin{array}{rcl}
1,  & {x_{ci}\in \xi^j}, \\
0,  & {x_{ci}\notin \xi^j},
\end{array} \right.
\end{equation}
is the coefficient to indicate whether point $x_{ci}$ is in quadrant
$\xi^j$ and $K_j$ is the number of KNN points in quadrant $\xi^j$.
Finally, all centroids of attributes ${\bf a}_{c}^{j}$, $j=1,\cdots,8$,
are concatenated to form a new descriptor of sampled point ${\bf p}_c$:
\begin{equation}\label{eq:attribute}
{\bf a}_c = Concat\{{\bf a}_{c}^{j}\}_{j=1}^{8}.
\end{equation}

This descriptor is robust with respect to disturbance in positions of
acquired points because of the averaging operation in each quadrant. We
use the 3D coordinates, $(x,y,z)$, as the initial attributes of a point.
It is called the 0-hop attributes. The dimension of 0-hop attributes is
3. The local descriptor as given in Eq. (\ref{eq:attribute}) has a
dimension of $3 \times 8=24$. We adopt the local descriptor as the new
attributes of a point that takes its relationship with its KNN neighbors
into account. It is called the 1-hop attributes. Note that the 0-hop
attributes can be generalized to $(x,y,z,r,g,b)$ for point clouds with
color information $(r,g,b)$ at each point. 

If ${\bf p}_B$ is a member in $KNN({\bf p}_A)$, we call that ${\bf p}_B$
is a 1-hop neighbor of ${\bf p}_A$. If ${\bf p}_C$ is a 1-hop neighbor
of ${\bf p}_B$ and ${\bf p}_B$ is a 1-hop neighbor of ${\bf p}_A$, we
call ${\bf p}_C$ is a 2-hop neighbor of ${\bf p}_A$ if ${\bf p}_C$ is
not a 1-hop neighbor of ${\bf p}_A$. The dimension of the attribute
vector of each point grows from 3 to 24 due to the change of local
descriptors from 0-hop to 1-hop. We can build another local descriptor
based on the 1-hop attributes of each point. The descriptor defines the
2-hop attributes of dimension $24 \times 8=192$. The $n$-hop attributes
characterize the relationship of a point with its $m$-hop neighbors, $m
\leq n$. 

As $n$ becomes larger, the $n$-hop attributes offer a larger coverage of
points in a point cloud model, which is analogous to a larger receptive
field in deeper layers of CNNs. Yet, the dimension growing rate is fast.
It is desired to reduce the dimension of the $n$-hop attribute vector
first before reaching out to neighbors of the $(n+1)$-hop. The Saab
transform \cite{kuo2019interpretable} is used to reduce the attribute
dimension of each point. A brief review of the Saab transform is given
in the Appendix. 

\begin{figure*}[htpb]
\centering
\centering
\subfigure[First unit]{\includegraphics[width=3in]{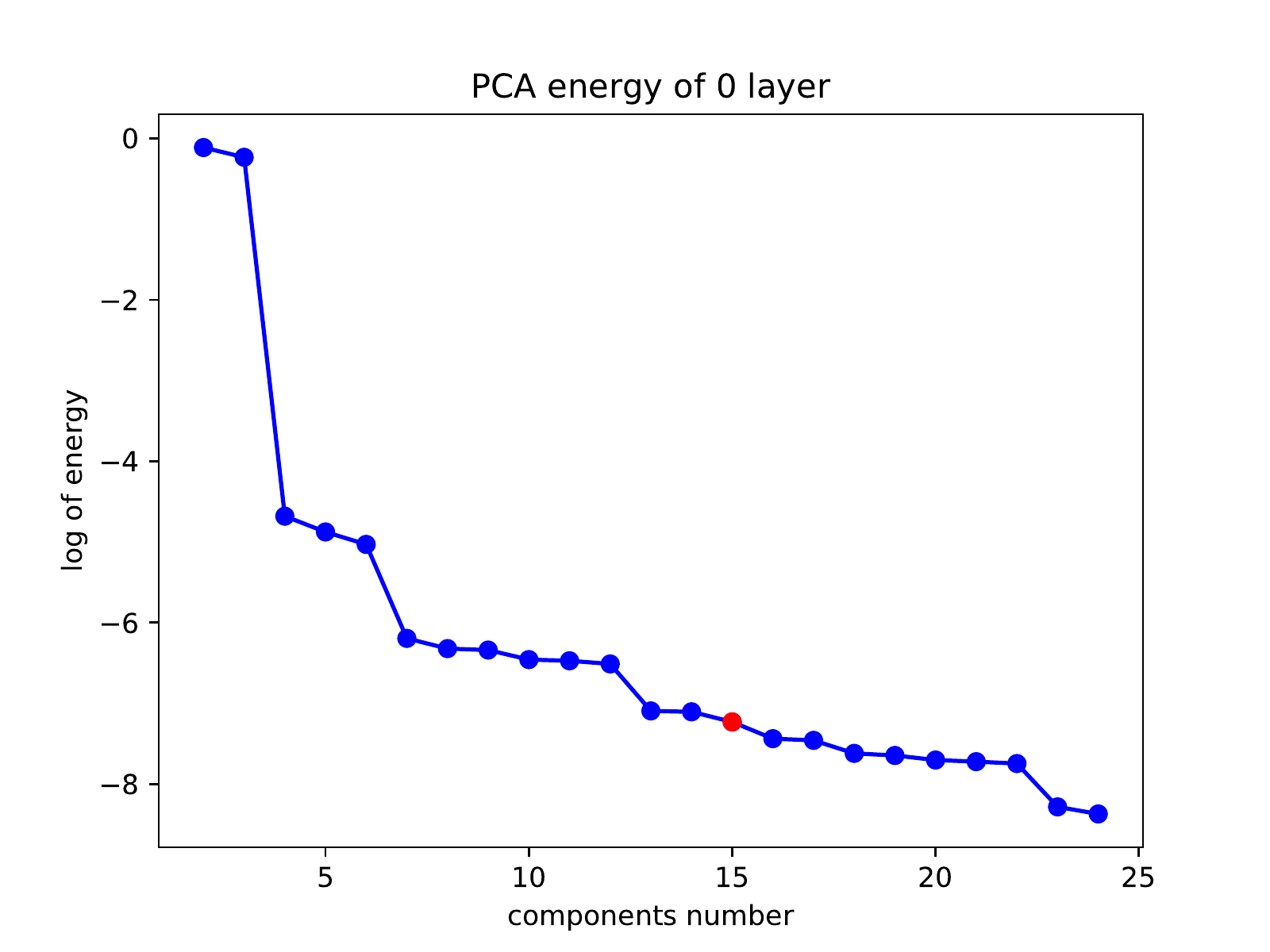}}
\subfigure[Second unit]{\includegraphics[width=3in]{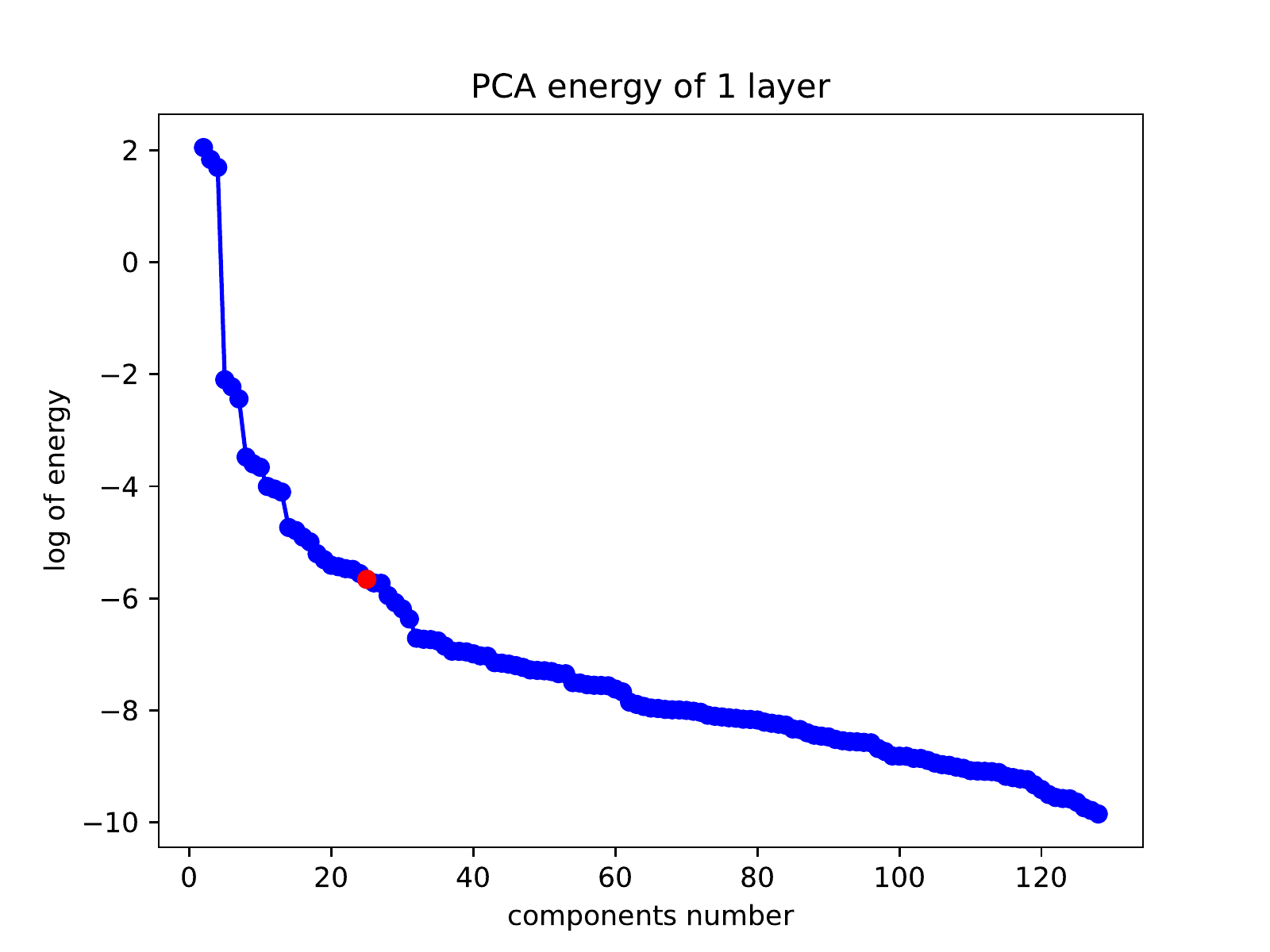}}
\subfigure[Third unit]{\includegraphics[width=3in]{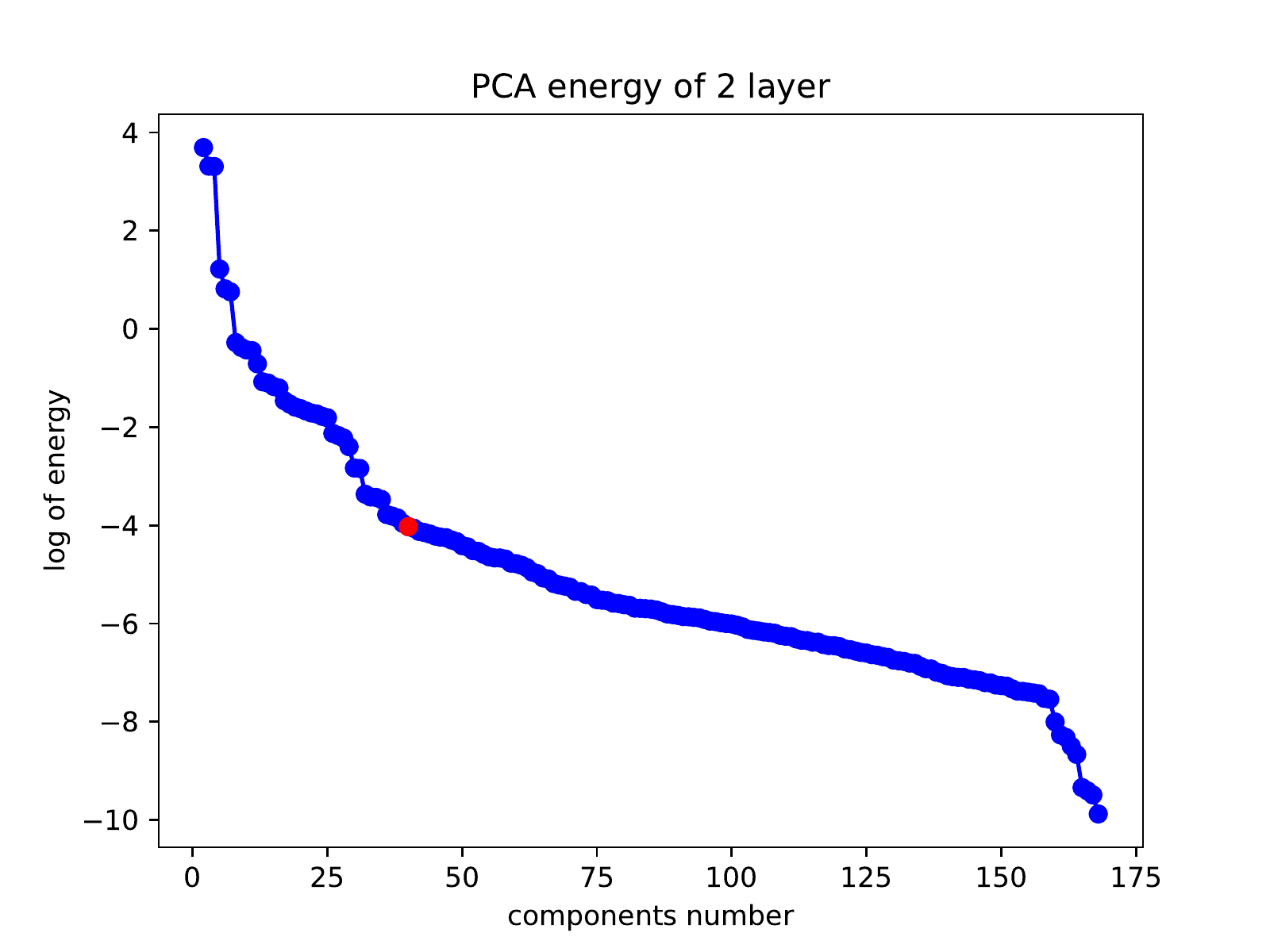}}
\subfigure[Fourth unit]{\includegraphics[width=3in]{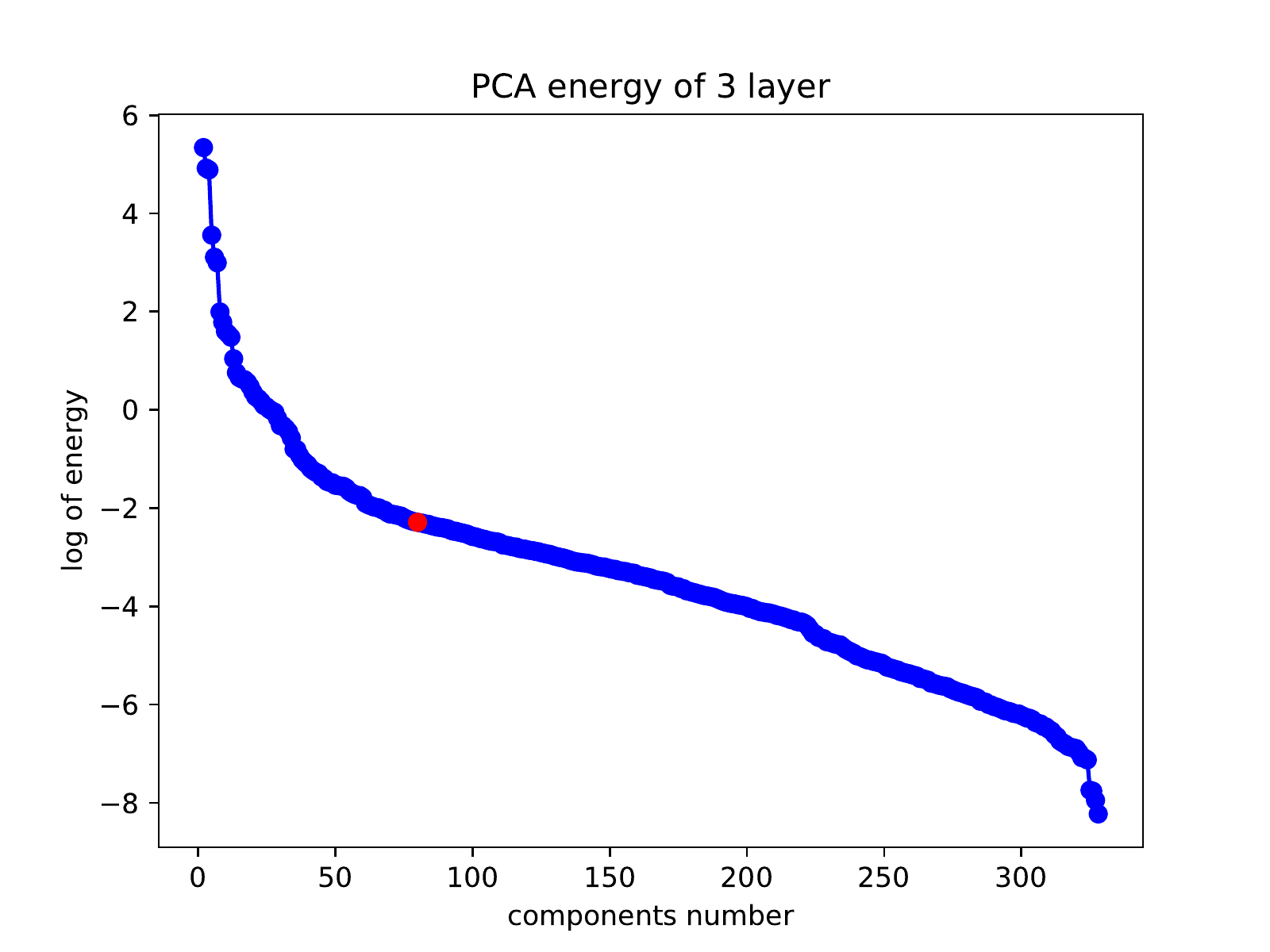}}
\caption{Determination of the number of Saab filters in each of the
PointHop units, where the red dot in each subfigure indicates the
selected number of Saab filters.} \label{fig:my_label-5}
\end{figure*}

Each PointHop unit has one-stage Saab transform. For $L$ PointHop units
in cascade, we need $L$-stage Saab transforms. We set $L=4$ in the
experiments. Each Saab transform contains three steps: 1) DC/AC
separation, 2) PCA and 3) bias addition. The number of AC Saab filters is determined by the energy plot of PCA coefficients as shown in Fig. \ref{fig:my_label-5}. We choose the knee location of the curve as indicated by the red point in each subfigure. 

The system diagram of the proposed PointHop method is shown in Fig.
\ref{fig:my_label-3}. It consists of multiple PointHop units. Four
PointHop units are shown in the figure. For the $i$th PointHop unit output, we use $N^i \times D^i$ to characterize its two parameters; namely, it has $N^i$ points and each of them has $D^i$ attributes. 

For the $i$th PointHop unit, we aggregate (or pool) each individual
attribute of $N^i$ points into a single feature vector. To enrich the
feature set, we consider multiple aggregation/pooling schemes such as
the max pooling \cite{qi2017pointnet}, the mean aggregation, the
$l_1$-norm aggregation and the $l_2$-norm aggregation. Then, we
concatenate them to obtain a feature vector of dimension $M \times
D^{i}$, where $M$ is the number of attribute aggregation methods, for
the $i$th PointHop unit. Finally, we concatenate feature vectors of all
PointHop units to form the ultimate feature vector of the whole system. 

To reduce computational complexity and speed up the coverage rate, we
adopt a spatial sampling scheme between two consecutive PointHop units
so that the number of points to be processed is reduced. This is
achieved by the farthest point sampling (FPS) scheme
\cite{katsavounidis1994new, eldar1997farthest, moenning2003fast} since
it captures the geometrical structure of a point cloud model better.
For a given set of input points, the FPS scheme first selects the point
closest to the centroid. Afterwards, it selects the point that has the
farthest Euclidean distance to existing points in the selected subset
iteratively until the target number is reached. The advantage of the FPS
scheme will be illustrated in Sec. \ref{sec:experiments}. 

\subsection{Classification and Ensembles}\label{subsec:decision}

Upon obtaining the feature vector, we adopt well known classifiers such
as the support vector machine (SVM) and the random forest (RF)
classifiers for the classification task. The SVM classifier performs
classification by finding gaps that separate different classes. Test
samples are then mapped into one of the side of the gap and predicted to
be the label of that side. The RF classifier first trains a number of
decision trees and each decision tree gives a output. Then, the RF
classifier ensembles outputs from all decision trees to give the mean
prediction. Both classifiers are mature and easy to use. 

Ensemble methods fuse results from multiple weak classifiers to get a
more powerful one \cite{dietterich2000ensemble, rokach2010ensemble, chen2019ensembles, zhang2012ensemble}. Ensembles are adopted in this
paper to improve the classification performance furthermore. We
consider the following two ensemble strategies. 
\begin{enumerate}
\item {\bf Decision ensemble.} Multiple PointHop units are individually
used as base classifiers and their decision vectors are concatenated to
form a new feature vector for the ensemble classifier. 
\item {\bf Feature ensemble.} Features from multiple PointHop units are
cascaded to form the final vector for the classification task. 
\end{enumerate}
It is our observation that the second strategy offers better
classification accuracy at the cost of a higher complexity if the
feature dimension is large. We choose the second strategy for its
higher accuracy. With the feature ensemble strategy, it is desired to
increase PointHop's diversity to enrich the feature set. We use the
following four schemes to achieve this goal. First, we augment the input
data by rotating it with a certain degree. Second, we change the number
of Saab filters in each PointHop unit. Third, we change the $K$ value in
the KNN scheme. Fourth, we vary the numbers of points in PointHop
units. 

\section{Experimental Results}\label{sec:experiments}

\begin{table*}[htpb]
    \centering
    \begin{tabular}{cc|cc|cccc|cc|cc|c} \hline
    \multicolumn{2}{c|}{Feature used} & \multicolumn{2}{|c|}{FPS} & \multicolumn{4}{|c|}{Pooling} & \multicolumn{2}{|c|}{Classifier} & \multicolumn{2}{|c|}{Dimension Reduction} & \multirow{2}*{Accuracy (\%)} \\ \cline{1-12} 
    All stages & Last stage & Yes & No & Max & Mean & $l_1$ & $l_2$ & SVM & Random Forest & PCA & Saab \\ \hline
    & \checkmark & \checkmark & & \checkmark & & & & \checkmark & & & \checkmark & 77.5 \\ 
    \checkmark & & & \checkmark & \checkmark & & & & \checkmark & & & \checkmark & 77.4 \\ 
    \checkmark & & \checkmark & & \checkmark & & & & \checkmark & & & \checkmark & 79.6 \\ 
    \checkmark & & \checkmark & & \checkmark & & & & \checkmark & & & \checkmark & 79.9 \\ 
    \checkmark & & & \checkmark & \checkmark & & & & & \checkmark & & \checkmark & 78.8 \\ 
    \checkmark & & \checkmark & & \checkmark & & & & & \checkmark & & \checkmark & 80.2 \\ 
    \checkmark & & \checkmark & & & \checkmark & & & & \checkmark & & \checkmark & 84.5 (default) \\
    \checkmark & & \checkmark & & & & \checkmark & & & \checkmark & & \checkmark & 84.8 \\
    \checkmark & & \checkmark & & & & & \checkmark & & \checkmark & & \checkmark & 85.6 \\
    \checkmark & & \checkmark & & \checkmark & \checkmark & & & & \checkmark & & \checkmark & 85.3 \\ 
    \checkmark & & \checkmark & & \checkmark & & \checkmark & & & \checkmark & & \checkmark & 85.7 \\
    \checkmark & & \checkmark & & \checkmark & & & \checkmark & & \checkmark & & \checkmark & 85.1 \\
    \checkmark & & \checkmark & & \checkmark & \checkmark & \checkmark & \checkmark & & \checkmark & & \checkmark & 86.1 \\    
    \checkmark & & \checkmark & & \checkmark & \checkmark & \checkmark & \checkmark & & \checkmark & \checkmark & & 85.6 \\    
    \hline 
    \end{tabular}
\caption{Results of ablation study with 256 sampled points as the 
input to the PointHop system.}\label{tab:my_label-1}
\end{table*}

We conduct experiments on a popular 3D object classification dataset
called ModelNet40 \cite{wu20153d}. The dataset contains 40 categories of
CAD models of objects such as airplanes, chairs, benches, cups, etc.
Each initial point cloud has 2,048 points and each point has three Cartesian
coordinates. There are 9,843 training samples and 2,468 testing samples. 

We adopt the following default setting in our experiments.
\begin{itemize}
    \item The number of sampled points into the first PointHop unit: 256 points.
    \item The sampling method from the input point cloud model to that
          as the input to the first PointHop unit: random sampling. 
    \item The number of $K$ in the KNN: $K=64$.
    \item The number of PointHop units in cascade: 4.
    \item The number of Saab AC filters in the $i$th PointHop unit: 15 ($i=1$), 25 ($i=2$), 
          40 ($i=3$) and 80 ($i=4$).
    \item The sampling method between PointHop units: Farthest Point Sampling (FPS).
    \item The number of sampled points in the 2nd, 3rd and 4th PointHop units: 128, 128 and 64.
    \item The aggregation method: mean pooling.
    \item The classifier: the random forest classifier.
    \item Ensembles: No.
\end{itemize}

This section is organized as follows. First, we conduct an ablation
study on an individual PointHop unit and show its robustness against the
sampling density variation in Sec. \ref{subsection41}. Next, we provide
results for various ensemble methods in Sec. \ref{subsection42}. Then,
we compare the performance of the proposed PointHop method and other
state-of-the-art methods in terms of accuracy and efficiency in Sec.
\ref{subsection43}. After that, we show activation maps of four layers
in Sec. \ref{subsection44}. Finally, we analyze hard samples in Sec.
\ref{subsection45}. 

\subsection{Ablation Study on PointHop Unit}\label{subsection41}

We show classification accuracy values under various parameter settings
in Table \ref{tab:my_label-1}. We see from the table that it is desired
to use features from all stages, the FPS between PointHop units,
ensembles of all pooling schemes, the random forest classifier and the Saab 
transform. As shown in the last row, we can reach a classification accuracy 
of 86.1\% with randomly selected 256 points as the input to the PointHop 
system. The whole training time is 5 minutes only. The FPS not only contributes
to higher accuracy but also reduces the computation time dramatically
since it can enlarge the receptive field in a faster rate. The RF
classifier has a higher accuracy than the SVM classifier. Besides, it
is much faster. 

\begin{figure}[htpb]
\centering
\includegraphics[width=3.5in]{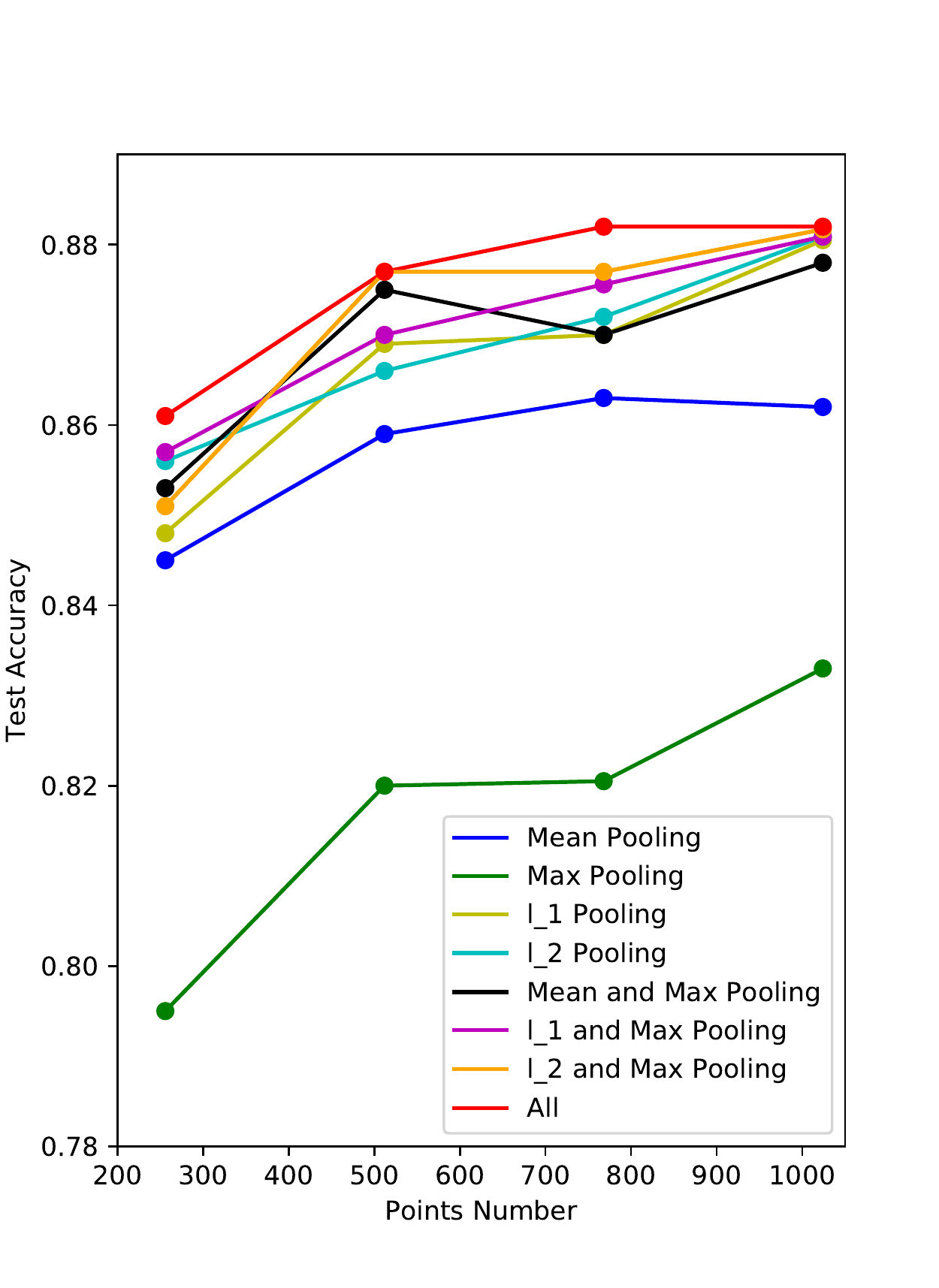}
\caption{The classification accuracy as a function of 
the sampled point number of the input model to the PointHop system
as well as different pooling methods.} \label{fig:my_label-6}
\end{figure}

\begin{table*}[!htpb]
    \centering
    \newcommand{\tabincell}[2]{\begin{tabular}{@{}#1@{}}#2\end{tabular}}
    \begin{tabular}{c|ccccc|c|c} \hline
     & Setting 1 & Setting 2 & Setting 3 & Setting 4 & Setting 5 & \multicolumn{2}{c}{Ensemble accuracy (\%)} \\ \hline
     HP-A & 0\degree & 45\degree & 90\degree & 135\degree & 180\degree & 88.0 & \multirow{4}*{88.0}\\ \cline{1-7} 
     HP-B & (15, 25, 40, 80) & (15, 25, 35, 50) & (18, 30, 50, 90) & (20, 40, 60, 100) & (20, 40, 70, 120) & 87.0 & \\ \cline{1-7} 
     HP-C & (64, 64, 64, 64) & (32, 32, 32, 32) & (32, 32, 64, 64) & (96, 96, 96, 96) & (128, 128, 128, 128) & 87.8 & \\ \cline{1-7} 
     HP-D & (512, 128, 128, 64) & (512, 256, 128, 64) & (512, 256, 256, 128) & (512, 256, 256, 256) & (512, 128, 128, 128) & 86.8 & \\ \hline
    \end{tabular}
\caption{Ensembles of five PointHops with changed hyper-parameter settings
and their corresponding classification accuracies.} \label{tab:my_label-2}
\end{table*}

We study the classification accuracy as a function of the sampled number
of all point cloud models as well as different pooling methods in Fig.
\ref{fig:my_label-6}, where the x-axis shows the number of sampled
points which is the same in training and testing. Corresponding to Fig.
\ref{fig:my_label-2}, we consider the following four settings: 256
points, 512 points, 768 points and 1,024 points. Different color curves
are obtained by different pooling schemes. We compare eight cases: four
individual ones, three ensembles of two, and one ensemble of all four.
We see that the maximum pooling and the mean pooling give the worst
performance. Their ensemble does not perform well, either. The
performance gap is small for the remaining five schemes as the point
number is 1,024. The ensemble of all pooling schemes given the best
results in all four cases. The highest accuracy is 88.2\% when we use
768 or 1,024 points with the ensemble of all four pooling schemes. 

\subsection{Ensembles of PointHop Systems}\label{subsection42}

Under the default setting, we consider ensemble five PointHops with changed
hyper-parameters (HP) to increase its diversity. They are summarized in
Table \ref{tab:my_label-2}. The hyper parameters of concern include the
following four. 
\begin{itemize}
\item {\bf HP-A.} We augment each point cloud model by rotating it 
with 45\degree four times.
\item {\bf HP-B.} We use different numbers of AC filters in the PointHop
units. 
\item {\bf HP-C.} We adopt different $K$ values in the KNN query in the
PointHop units. 
\item {\bf HP-D.} We take point cloud models of different point numbers 
as the input to the PointHop units in four stages.
\end{itemize}

For HP-B, HP-C and HP-D, the four numbers in the table correspond to
those in the first, second, third and fourth PointHop units,
respectively. To get ensemble results of HP-A, we keep HP-B, HP-C and
HP-D the same (say, Setting 1). The same procedure applies in getting
the ensemble results of HP-B, HP-C and HP-D. Furthermore, we can derive
ensemble results of all cases as shown in the last column. We see from
the table that the most simple and effective ensemble result is achieved
by rotating point clouds, where we can reach the test accuracy of 88\%.
Thus, we focus on this ensemble method only in later experiments.

\subsection{Comparison with State-of-the-Art Methods}\label{subsection43}

We first compare the classification accuracy of the proposed PointHop
system with those of several state-of-the-art methods such as PointNet
\cite{qi2017pointnet}, PointNet++ \cite{qi2017pointnet++}, PointCNN
\cite{li2018pointcnn} and DGCNN \cite{wang2018dynamic} in Table
\ref{tab:my_label-3}. All of these works (including ours) are based on
the model of 1,024 points. The column of ``average accuracy" means the
average of per-class classification accuracy while the column of
``overall accuracy" shows the best result obtained. Our PointHop
baseline containing a single model without any ensembles can achieve
88.65\% overall accuracy. With ensemble, the overall accuracy is
increased to 89.1\%. The performance of PointHop is worse than that of
PointNet \cite{li2018pointcnn} and DGCNN \cite{wang2018dynamic} by 0.1\%
and 3.1\%, respectively. On the other hand, our PointHop method
performs better than other unsupervised methods such as LFD-GAN
\cite{achlioptas2017representation} and FoldingNet
\cite{yang2018foldingnet}.

\begin{table}[!htpb]
    \centering
    \newcommand{\tabincell}[2]{\begin{tabular}{@{}#1@{}}#2\end{tabular}}
    \begin{tabular}{c|c|c|c} \hline
    Method & \tabincell{c}{Feature \\ extraction} & \tabincell{c}{Average \\ accuracy (\%)} & \tabincell{c}{Overall \\ accuracy (\%)}  \\ \hline
    PointNet \cite{qi2017pointnet} & \multirow{4}*{Supervised} & 86.2 & 89.2 \\
    PointNet++ \cite{qi2017pointnet++} & & - & 90.7 \\
    PointCNN \cite{li2018pointcnn} & & 88.1 & 92.2 \\
    DGCNN \cite{wang2018dynamic} & & 90.2 &92.2 \\ \hline
    \tabincell{c}{PointNet baseline \\ (Handcraft, MLP)} & \multirow{6}*{Unsupervised} & 72.6 & 77.4 \\
    LFD-GAN \cite{achlioptas2017representation} & & - & 85.7 \\
    FoldingNet \cite{yang2018foldingnet} & & - & 88.4 \\
    PointHop (baseline) & & 83.3 & 88.65 \\ 
    PointHop & & 84.4 & {\bf 89.1} \\ \hline
    \end{tabular}
\caption{Comparison of classification accuracy on ModelNet40, where the
proposed PointHop system achieves 89.1\% test accuracy, which is 0.1\% less
than PointNet \cite{qi2017pointnet} and 3.1\% less than DGCNN
\cite{wang2018dynamic}.} \label{tab:my_label-3}
\end{table}

Next, we compare the time complexity in Table \ref{tab:my_label-4}. As
shown in the table, the training time of the PointHop system is
significantly lower than deep-learning-based methods. It takes 5 minutes
and 20 minutes in training a PointHop baseline of 256-point and
1,024-point cloud models, respectively, with CPU. Our CPU is Intel(R)
Xeon(R) CPU E5-2620 v3 at 2.40GHz. In contrast, PointNet
\cite{qi2017pointnet} takes more than 5 hours in training using one
GTX1080 GPU. Furthermore, we compare the inference time in the test
stage.  PointNet++ demands 163.2ms in classifying a test sample of 1024
points while our PointHop method only needs 108.4ms. The most time
consuming module in the PointHop system is the KNN query that compares
the distance between points. It is possible to lower training/testing
time even more by speeding up this module. 

\begin{table}[!htpb]
\centering
\newcommand{\tabincell}[2]{\begin{tabular}{@{}#1@{}}#2\end{tabular}}
\begin{tabular}{c|ccc} \hline
Method & \tabincell{c}{Total \\ training time} & \tabincell{c}{Inference \\ time (ms)} & Device  \\ \hline
PointNet (1,024 points) & $\sim$ 5 hours & 25.3 & GPU \\ 
PointNet++ (1,024 points) & - & 163.2 & GPU \\ 
PointHop (256 points) & $\sim$ 5 minutes & 103 & CPU \\ 
PointHop (1,024 points) & $\sim$ 20 minutes & 108.4 & CPU \\ \hline
\end{tabular}
\caption{Comparison of time complexity between PointNet/PointNet++ and
PointHop.}\label{tab:my_label-4}
\end{table}

\begin{figure}[htpb]
\centering
\includegraphics[width=3.5in]{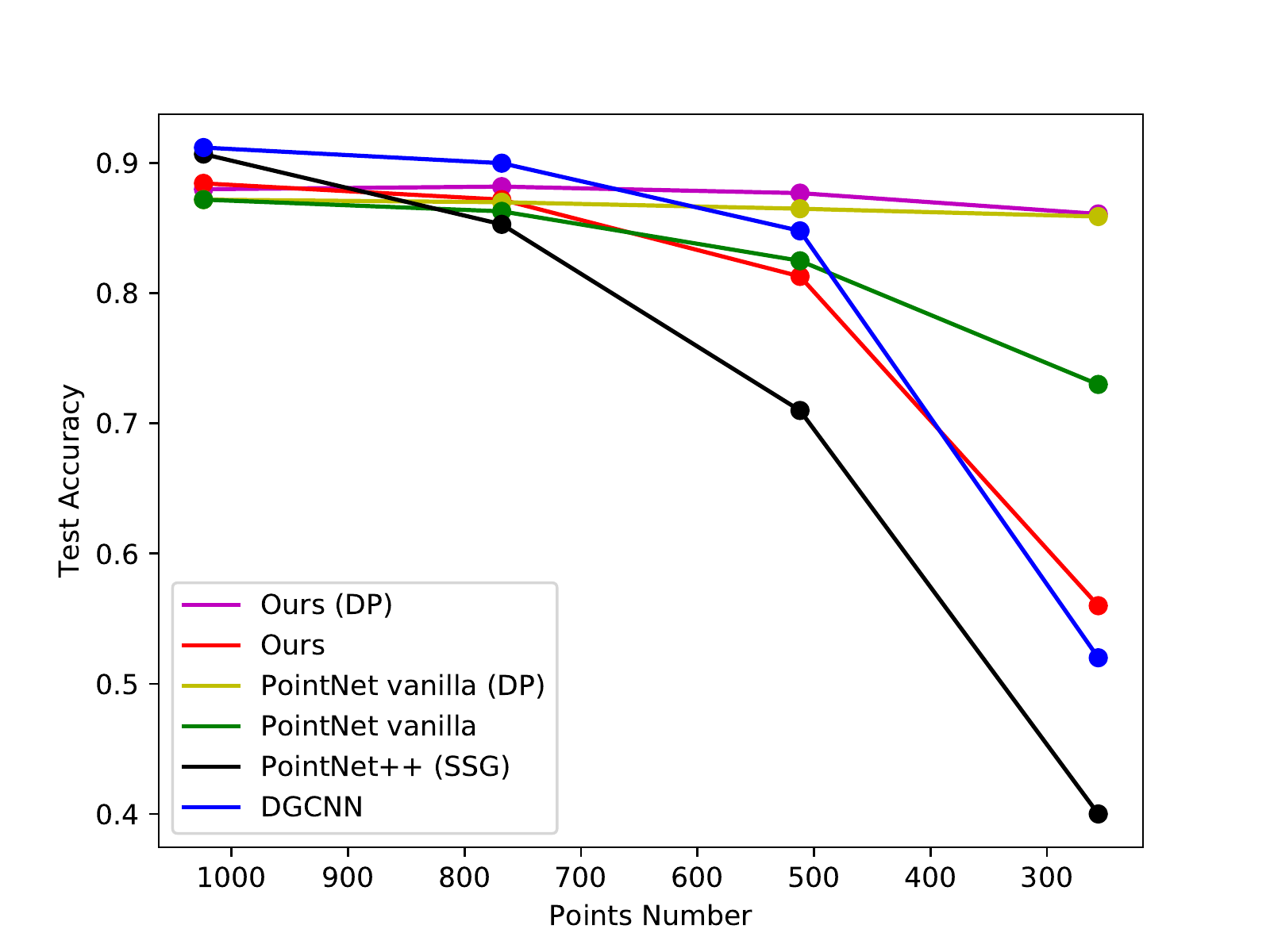}
\caption{Robustness to sampling density variation: comparison of test
accuracy as a function of sampled point numbers of different
methods.}\label{fig:my_label-7}
\end{figure}

In Fig. \ref{fig:my_label-7}, we examine the robustness of
classification performance with respect to models of four point numbers,
i.e., 256, 512, 768 and 1,024. For the first scenario, the numbers in
training and testing are the same. It is indicated by DP in the legend.
The PointHop method and the PointNet vanilla are shown in violet and
yellow lines. The PointHop method with DP is more robust than PointNet
vanilla with DP. For the second scenario, we train each method based on
1,024-point models and, then, apply the trained model to point clouds of
the same or fewer point numbers in the test. For the latter, there is a
point cloud model mismatch between training and testing. We see that the
PointHop method is more robust than PointNet++ (SSG) in the mismatched
condition. The PointHop method also outperforms DGCNN in the mismatched
condition of the 256-point models. 

\subsection{Feature Visualization}\label{subsection44}

\begin{figure*}
\centering
\includegraphics[width=6.5in]{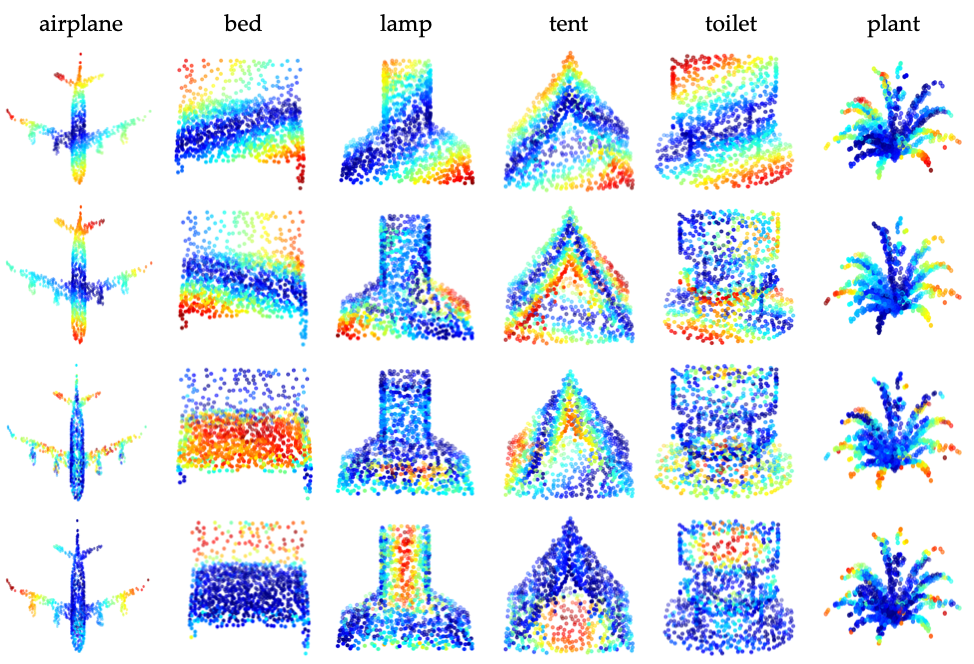}
\caption{Visualization of learned features in the first-stage PointHop
unit.}\label{fig:my_label-8}
\end{figure*}

The learned features of the first-stage PointHop Unit are visualized in
Fig. \ref{fig:my_label-8} for six highly varying point cloud models. We
show the responses of different channels that are normalized into
$[0,1]$ (or from blue to red in color). We see that many common patterns
are learned such as corners of tents/lamps and plans of airplanes/beds.
The learned features comprise powerful and informative description in
the 3D geometric space. 

\subsection{Error Analysis}\label{subsection45}

The average accuracy of the PointHop method is worse than PointNet
\cite{qi2017pointnet} by 1.8\%. To provide more insights, we show
per-class accuracy on ModelNet40 in Table \ref{tab:my_label-5}. We see
that PointHop achieves equal or higher accuracy in 18 classes. On the
other hand, it has low accuracy in several classes, including flower-pot
(10\%), cup (55\%), radio (65\%) and sink (60\%). Among them, the
flower pot is the most challenging one. 

\begin{table*}[!htbp]
    \centering
    \begin{tabular}{c|cccccccccc} \hline
    Network & airplane & bathtub & bed & bench & bookshelf & bottle & bowl & car & chair & cone \\ \hline
    PointNet & 100.0 & 80.0 & 94.0 & 75.0 & 93.0 & 94.0 & 100.0 & 97.9 & 96.0 & 100.0 \\
    PointHop & 100.0 & 94.0 & 99.0 & 70.0 & 96.0 & 95.0 & 95.0 & 97.0 & 100.0 & 90.0 \\ \hline
         & cup & curtain & desk & door & dresser & flower pot & glass box & guitar & keyboard & lamp \\ \hline
    PointNet & 70.0 & 90.0 & 79.0 & 95.0 & 65.1 & 30.0 & 94.0 & 100.0 & 100.0 & 90.0 \\
    PointHop & 55.0 & 85.0 & 90.7 & 90.0 & 83.7 & 10.0 & 95.0 & 99.0 & 95.0 & 75.0 \\ \hline    
         & laptop & mantel & monitor & night stand & person & piano & plant & radio & range hood & sink \\ \hline
    PointNet & 100.0 & 96.0 & 95.0 & 82.6 & 85.0 & 88.8 & 73.0 & 70.0 & 91.0 & 80.0 \\
    PointHop & 100.0 & 91.0 & 98.0 & 79.1 & 80.0 & 82.0 & 76.0 & 65.0 & 91.0 & 60.0 \\ \hline  
         & sofa & stairs & stool & table & tent & toilet & tv stand & vase & wardrobe & xbox \\ \hline
    PointNet & 96.0 & 85.0 & 90.0 & 88.0 & 95.0 & 99.0 & 87.0 & 78.8 & 60.0 & 70.0 \\
    PointHop & 96.0 & 75.0 & 85.0 & 82.0 & 95.0 & 97.0 & 82.0 & 84.0 & 70.0 & 75.0 \\ \hline         
    \end{tabular}
\caption{Comparison of per-class classification accuracy on the ModelNet40.}\label{tab:my_label-5}
\end{table*}

\begin{figure*}[htpb]
\centering
\subfigure[flower pot test samples]{\includegraphics[width=3.5in]{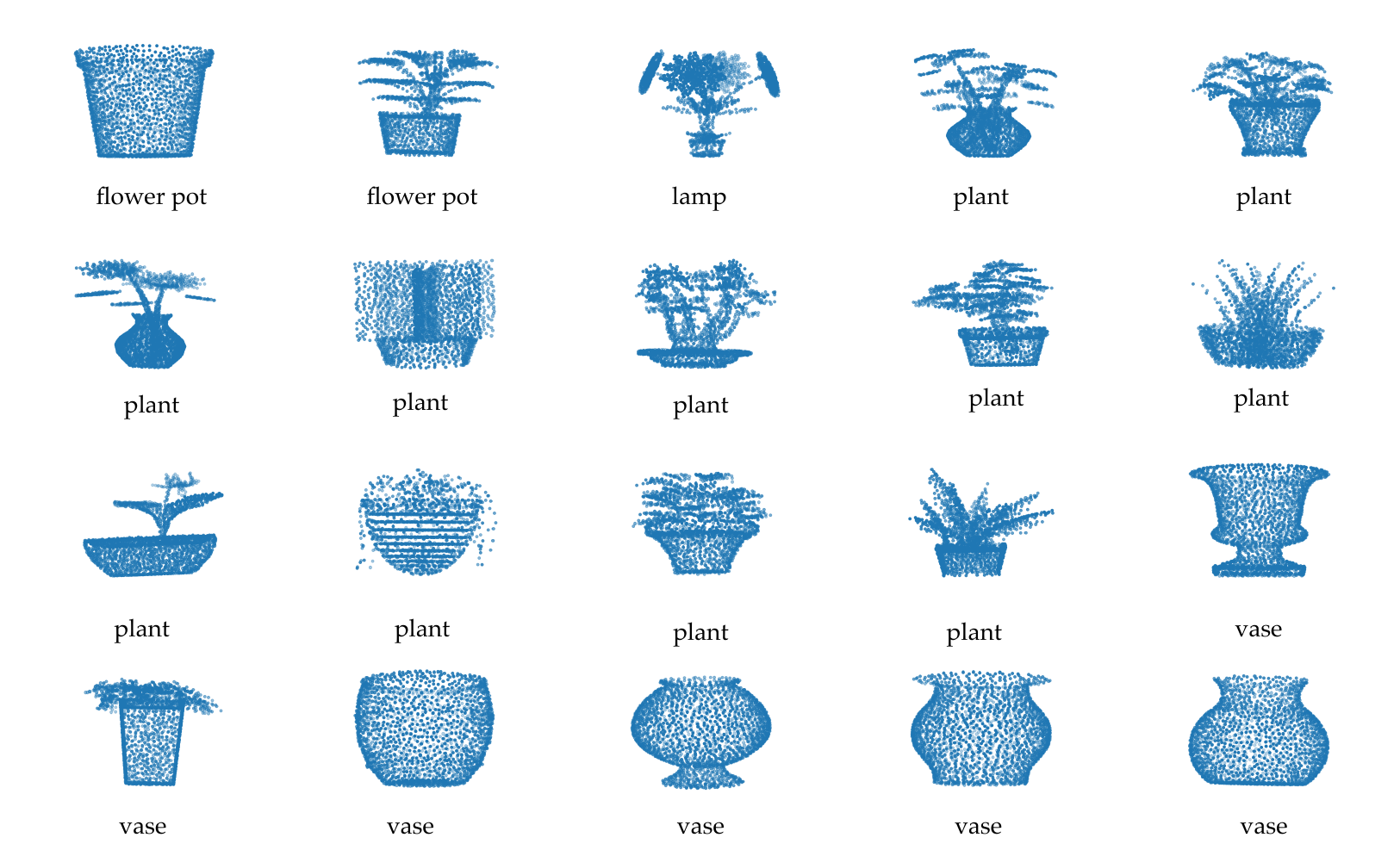}}
\subfigure[cup test samples]{\includegraphics[width=3.5in]{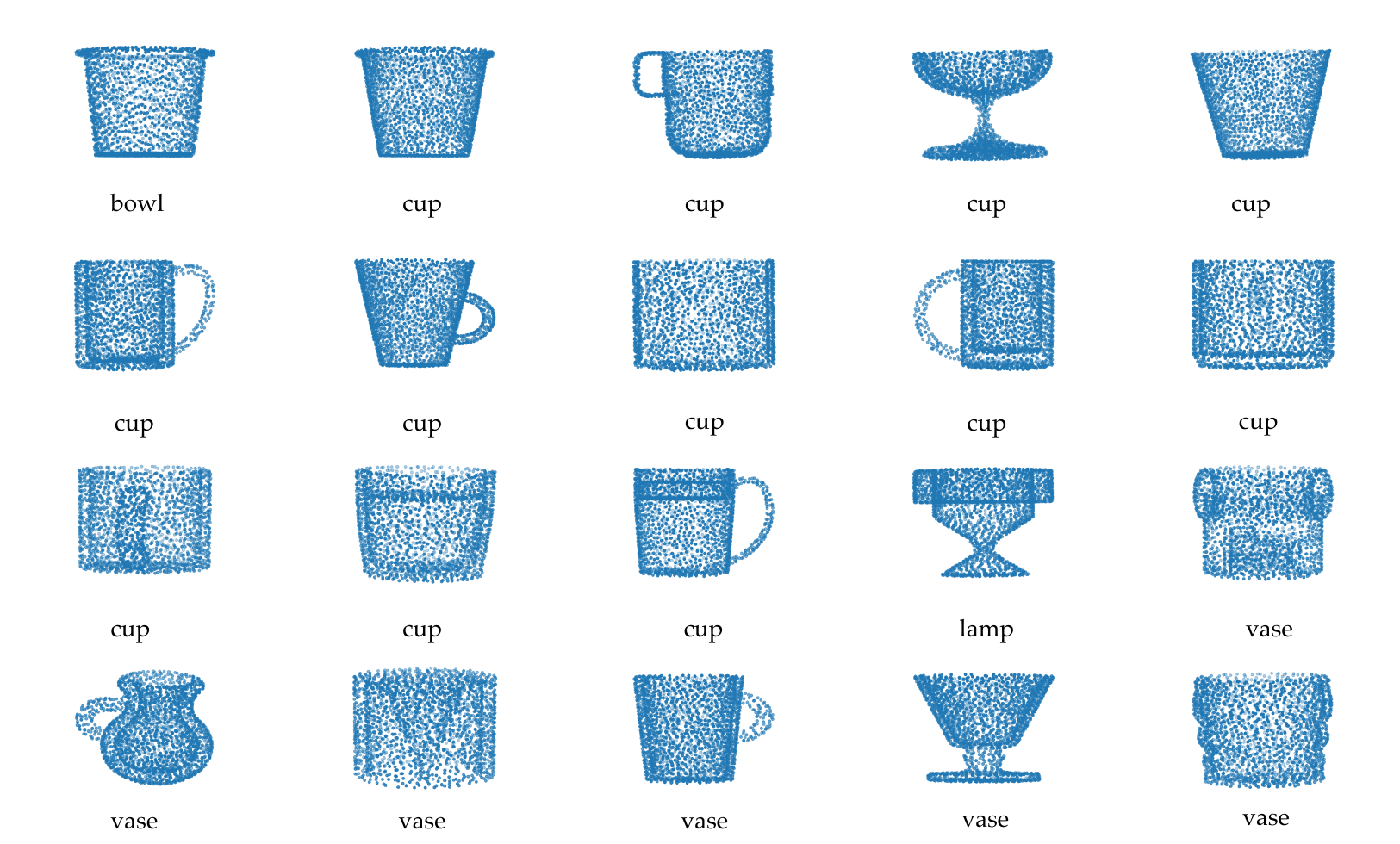}}
\caption{The label under each point cloud is its predicted class. Many
flower pots are misclassified to the plant and the vase classes. Also,
quite a few cups are misclassified to the vase class.}\label{fig:my_label-9}
\end{figure*}

We conduct error analysis on two object classes, ``flower pot" and
``cup", in Figs. \ref{fig:my_label-9} (a) and (b), respectively. The
total test number of the flower pot class is 20. Eleven, six and one of
them are misclassified to the plant, the vase and the lamp classes,
respectively. There are only two correct classification cases. We show
all point clouds of the flower pot class in Fig. \ref{fig:my_label-9}
(a). Only the first point cloud has a unique flower pot shape while
others have both the flower pot and the plant or are similar to the vase
in shape. As to the cup class classification, six are misclassified to
the vase class, one misclassified to the bowl class and another one
misclassified to the lamp class. There are twelve correct classification
results. The errors are caused by shape/functional similarity. To
overcome the challenge, we may need to supplement the data-driven approach
with the rule-based approach to improve the classification performance
furthermore. For example, the height-to-radius ratio of a flower pot is
smaller than that of a vase. Also, if the object has a holder, it is
more likely to be a cup rather than a vase. 

\section{Conclusion}\label{sec:conclusion}

An explainable machine learning method called the PointHop method was
proposed for point cloud classification in this work. It builds
attributes of higher dimensions at each sampled point through iterative
one-hop information exchange. This is analogous to a larger receptive
field in deeper convolutional layers in CNNs. The problem of unordered
point cloud data was addressed using a novel space partitioning
procedure. Furthermore, we used the Saab transform to reduce the
attribute dimension in each PointHop unit. In the classification stage,
we fed the feature vector to a classifier and explored ensemble methods
to improve the classification performance. It was shown by experimental
results that the training complexity of the PointHop method is
significantly lower than that of state-of-the-art deep-learning-based
methods with comparable classification performance. We conducted error
analysis on hard object classes and pointed out a future research
direction for further performance improvement by considering data-driven
and rule-based approaches jointly. 

\section*{Appendix: Saab Transform}\label{sec:appendix}

The principal component analysis (PCA) is a commonly used
dimension reduction technique. The Saab transform uses a specific way to
conduct multi-stage PCAs. For an input ${\bf v}=(v_0,v_1,\cdots,
v_{N-1})^T$ of dimension $N$, the one-stage Saab transform can be
written as
\begin{equation}
y_k=\sum_{n=0}^Na_{k,n}v_n+b_k={{\bf a}^{T}_{k}}{\bf v}+b_k, \quad k=0, 
\cdots, K-1   
\end{equation}
where $y_k$ is the $k$th Saab coefficient, ${\bf a}_{k}=(a_{k,0},
a_{k,1}, \cdots, a_{k,N-1})^T$ is the weight vector and $b_k$ is the
bias term for the $k$th Saab filter. The Saab transform has a
particular rule in choosing filter weight ${\bf a}_k$ and bias term
$b_k$. 

Let us focus on filter weights first. When $k=0$, the filter is called
the DC (direct current) filter, and its filter weight is
$$
{\bf a}_0={\frac{1}{\sqrt{N}}(1,\cdots,1)^T}.
$$ 
By projecting input ${\bf v}$ to the DC filter, we get its DC component 
${\bf v}_{DC}={\frac{1}{\sqrt{N}}{\sum_{n=0}^N}v_n}$, which is nothing
but the local mean of the input. We can derive the AC component of the 
input via
$$
{\bf v}_{AC}={\bf v}-{\bf v}_{DC}. 
$$
When $k>0$, the filters are called the AC (alternating current) filters.
To derive AC filters, we conduct PCA on AC components, ${\bf v}_{AC}$,
and choose its first $(K-1)$ principle components as the AC filters
${\bf a}_k, k=1,\cdots,K-1$. Finally, the DC filter and $K-1$ AC filters
form the set of Saab filters.

Next, we discuss the choice of the bias term, $b_k$, of the $k$th
filter. In CNNs, there is an activation function at the output of each
convolutional operation such as the ReLU (Rectified Linear Unit) and the
sigmoid. In the Saab transform, we demand that all bias terms are the
same so that they contribute to the DC term in the next stage. Besides,
we choose the bias large enough to guarantee that the response $y_k$ is
always non-negative before the nonlinear activation operation. Thus,
nonlinear activation plays no role and can be removed. It is
shown in \cite{kuo2019interpretable} that $b_k$ can be selected using
the following rule:
$$
b_k = \mbox{constant} \geq{\max\limits_{\bf v}{\left \| {\bf v} 
\right \|}}, \quad k=0, \cdots, N-1.
$$

Pixels in images have a decaying correlation structure. The correlation
between local pixels is stronger and the correlation becomes weaker as
their distance becomes larger. To exploit this property, we conduct the
first-stage PCA in a local window for dimension reduction to get a local
spectral vector. It will result in a joint spatial-spectral cuboid
where the spatial dimension denotes the spatial location of the local
window and the spectral dimension provides the spectral components of
the corresponding window. Then, we can perform the second-stage PCA on
the joint spatial-spectral cuboid. The multi-stage PCA is better than
the single-stage PCA since it handles decaying spatial correlations in
multiple spatial resolutions rather than in a single spatial resolution. 

\section*{Acknowledgement}\label{sec:acknowledgement}

This work was supported by a research grant from Tencent.

\ifCLASSOPTIONcaptionsoff
  \newpage
\fi

\bibliographystyle{IEEEtran}
\bibliography{main}

\begin{IEEEbiography}
[{\includegraphics[width=1in,height=1.25in,clip,keepaspectratio]{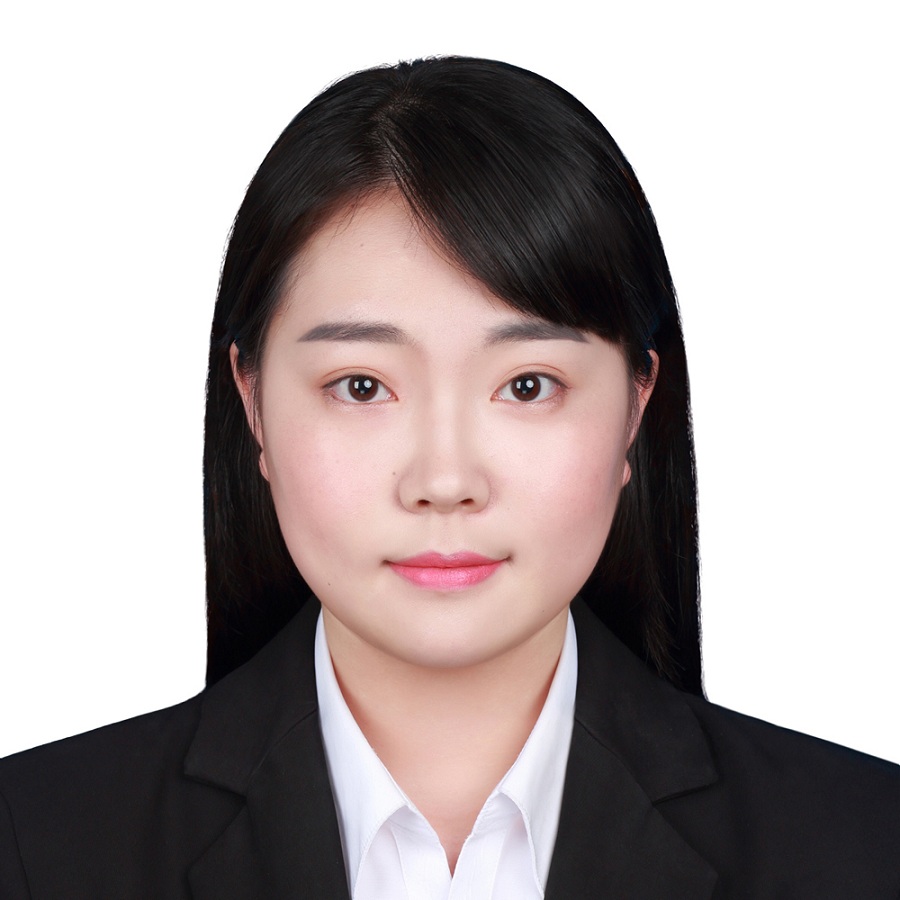}}]
{Min Zhang} received her B.E. degree from the School of Science, Nanjing
University of Science and Technology, Nanjing, China and her M.S. degree
from the Viterbi School of Engineering, University of Southern
California, Los Angeles, US, in 2017 and 2019, respectively. She is
currently working toward the Ph.D. degree from University of Southern
California. Her research interests include pattern recognition and
machine learning, image and video processing, object segmentation,
detection and tracking. 
\end{IEEEbiography}

\begin{IEEEbiography}
[{\includegraphics[width=1in,height=1.25in,clip,keepaspectratio]{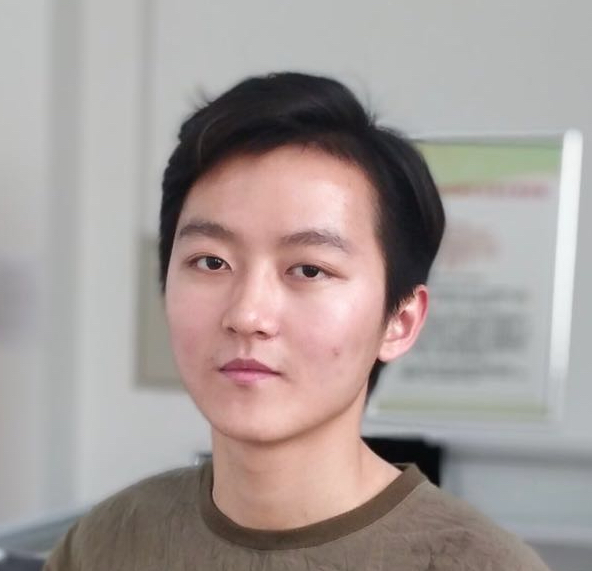}}]
{Haoxuan You} received his Bachelor of Engineering degree in Electronics
Information Engineering from Xidian University, Xian, China in 2018. He
is currently pursuing Ph.D. degree in Computer Science from Columbia
University, New York, USA. His research interests lie in computer vision
including 3D object recognition, 3D object detection, multimodal
learning and machine learning especially hypergraph learning. 
\end{IEEEbiography}

\begin{IEEEbiography}
[{\includegraphics[width=1in,height=1.25in,clip,keepaspectratio]{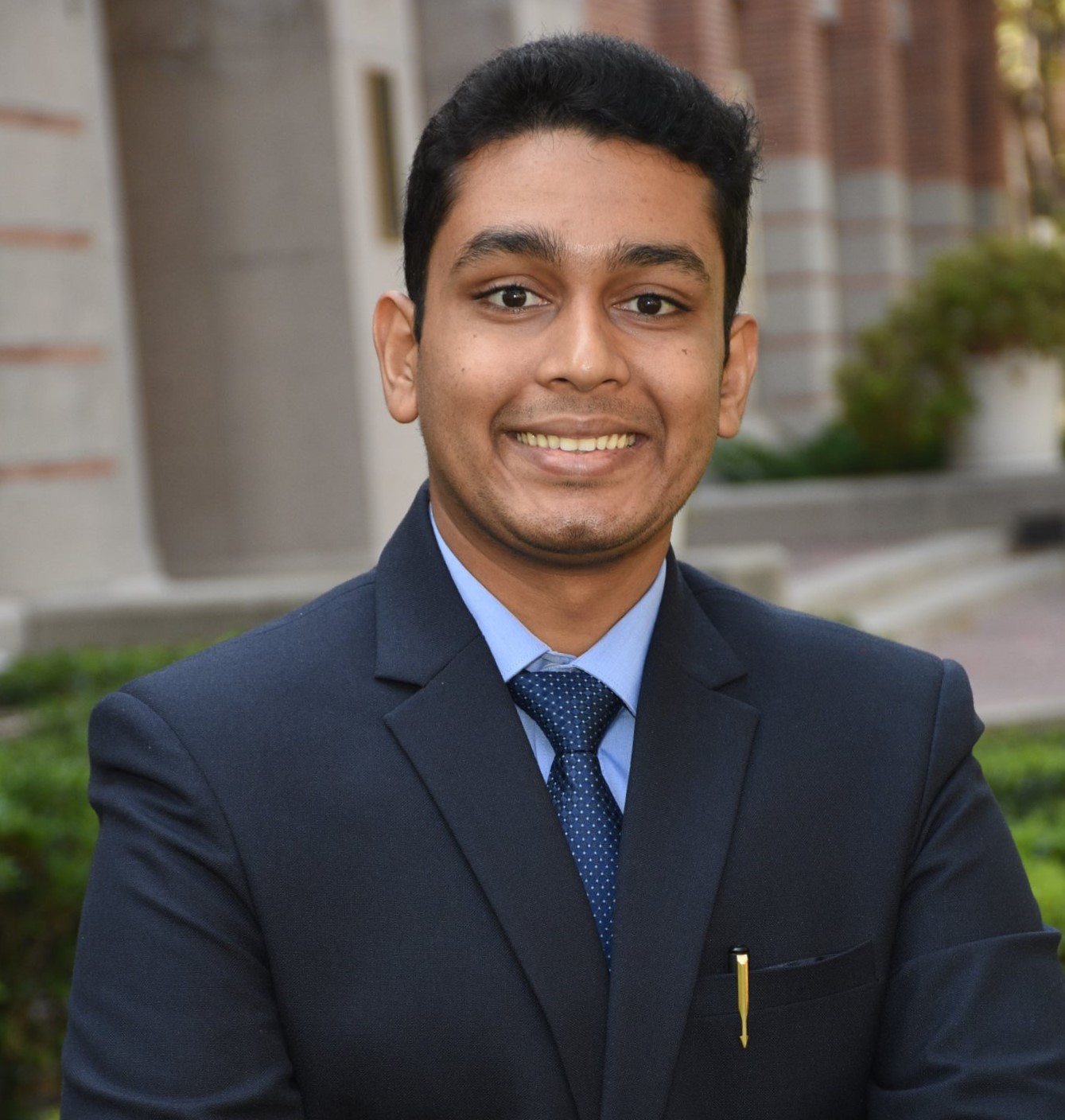}}]
{Pranav Kadam} received the Bachelor of Engineering degree in
Electronics and Telecommunication from Savitribai Phule Pune University,
Pune, India in 2018. He is currently pursuing M.S. degree in Electrical
Engineering with specialization in signal and image processing from
University of Southern California, Los Angeles, USA. His research
interests include computer vision and applications of machine learning
and deep learning techniques in image and video analysis. 
\end{IEEEbiography}

\begin{IEEEbiography}
[{\includegraphics[width=1in,height=1.25in,clip,keepaspectratio]{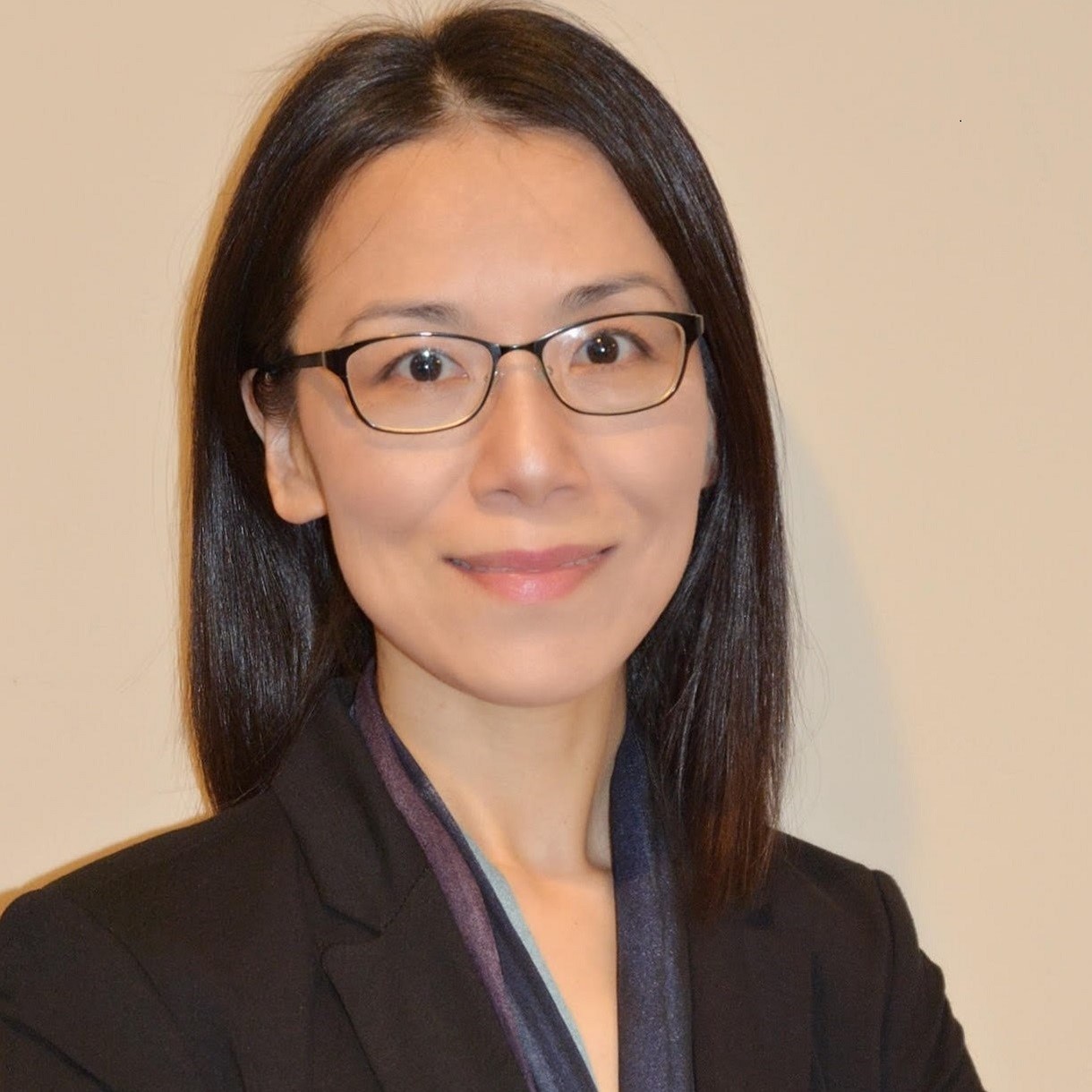}}]
{Shan Liu} is a Distinguished Scientist and General Manager at Tencent where she heads the Tencent Media Lab. Prior to joining Tencent she was the Chief Scientist and Head of America Media Lab at Futurewei Technologies. She was formerly Director of Multimedia Technology Division at MediaTek USA. She was also formerly with MERL, Sony and IBM. Dr. Liu is the inventor of more than 200 US and global patent applications and the author of more than 50 journal and conference articles. She actively contributes to international standards such as VVC, H.265/HEVC, DASH, OMAF, and served as co-Editor of H.265/HEVC v4 and VVC. Dr. Liu obtained her B.Eng. degree in Electronics Engineering from Tsinghua University, Beijing, China and M.S. and Ph.D. degrees in Electrical Engineering from University of Southern California, Los Angeles, USA.

\end{IEEEbiography}
\begin{IEEEbiography}
[{\includegraphics[width=1in,height=1.25in,clip,keepaspectratio]{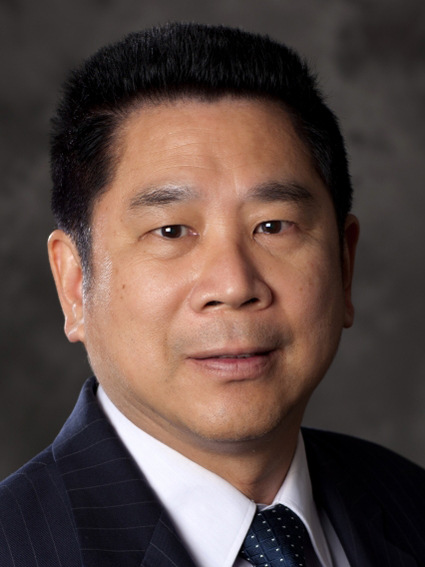}}]
{C.-C. Jay Kuo}(F'99) received the B.S. degree in electrical
engineering from the National Taiwan University, Taipei, Taiwan, in
1980, and the M.S. and Ph.D. degrees in electrical engineering from the
Massachusetts Institute of Technology, Cambridge, in 1985 and 1987,
respectively. He is currently the Director of the Multimedia
Communications Laboratory and a Distinguished Professor of electrical
engineering and computer science at the University of Southern
California, Los Angeles. His research interests include digital
image/video analysis and modeling, multimedia data compression,
communication and networking, and biological signal/image processing. He
is the coauthor of about 280 journal papers, 940 conference papers and
14 books. Dr. Kuo is a Fellow of the American Association for the
Advancement of Science (AAAS) and The International Society for Optical
Engineers (SPIE). 
\end{IEEEbiography}

\end{document}